%% file: main.tex
\documentclass[letterpaper, 10 pt, conference]{ieeeconf}  % Comment this line out
\IEEEoverridecommandlockouts                              % This command is only
\input{preamble_and_macros}

\title{\LARGE \bf On Controller Tuning with Time-Varying Bayesian Optimization}

\author{Paul Brunzema\textsuperscript{1 *}, Alexander von Rohr\textsuperscript{1,2,3 *}, Sebastian Trimpe\textsuperscript{1,2}% <-this % stops a space% <-this % stops a space
\thanks{\raggedright\textsuperscript{1}Institute for Data Science in {Mechanical Engineering}, RWTH Aachen University,
        Aachen, Germany, \{firstname.lastname\}@dsme.rwth-aachen.de}
\thanks{\noindent\textsuperscript{2}Max Planck Institute for Intelligent Systems, Stuttgart, Germany}
\thanks{\noindent\textsuperscript{3}IAV GmbH Ingenieurgesellschaft Auto und Verkehr, Germany}
\thanks{\noindent*The authors contributed equally.}%
\thanks{This work was supported in part by the Cyber Valley Initiative and the Max Planck Society.
The authors thank the International Max Planck Research School for Intelligent Systems for supporting A.~von~Rohr.}
}

\input{glossary}

\glsdisablehyper

\begin{document}

\maketitle
% \copyrightnotice
\thispagestyle{fancy}
\pagestyle{empty}

%%%%%%%%%%%%%%%%%%%%%%%%%%%%%%%%%%%%%%%%%%%%%%%%%%%%%%%%%%%%%%%%%%%%%%%%%%%%%%%%
\input{moin_matter/abstract}

%%%%%%%%%%%%%%%%%%%%%%%%%%%%%%%%%%%%%%%%%%%%%%%%%%%%%%%%%%%%%%%%%%%%%%%%%%%%%%%%
\input{moin_matter/introduction}

%%%%%%%%%%%%%%%%%%%%%%%%%%%%%%%%%%%%%%%%%%%%%%%%%%%%%%%%%%%%%%%%%%%%%%%%%%%%%%%%
\input{moin_matter/related_work}

%%%%%%%%%%%%%%%%%%%%%%%%%%%%%%%%%%%%%%%%%%%%%%%%%%%%%%%%%%%%%%%%%%%%%%%%%%%%%%%%
\input{moin_matter/methods}

%%%%%%%%%%%%%%%%%%%%%%%%%%%%%%%%%%%%%%%%%%%%%%%%%%%%%%%%%%%%%%%%%%%%%%%%%%%%%%%%
\input{moin_matter/results}

%%%%%%%%%%%%%%%%%%%%%%%%%%%%%%%%%%%%%%%%%%%%%%%%%%%%%%%%%%%%%%%%%%%%%%%%%%%%%%%%

% \addtolength{\textheight}{-3cm}   % This command serves to balance the column lengths

%%%%%%%%%%%%%%%%%%%%%%%%%%%%%%%%%%%%%%%%%%%%%%%%%%%%%%%%%%%%%%%%%%%%%%%%%%%%%%%%
\input{moin_matter/conclussion}

%%%%%%%%%%%%%%%%%%%%%%%%%%%%%%%%%%%%%%%%%%%%%%%%%%%%%%%%%%%%%%%%%%%%%%%%%%%%%%%%
\section*{Acknowledgments}

% Danke
The authors thank D.~Baumann, C.~Fiedler, F.~Hübenthal, P.-F.~Massiani, and D.~Stenger for their helpful comments and discussions on this paper and TVBO.

%%%%%%%%%%%%%%%%%%%%%%%%%%%%%%%%%%%%%%%%%%%%%%%%%%%%%%%%%%%%%%%%%%%%%%%%%%%%%%%%
% \clearpage
% \newpage
\printbibliography
\end{document}

%% file: preamble_and_macros.tex
\usepackage{amsthm}
\usepackage[hidelinks]{hyperref}
\usepackage[utf8]{inputenc}
\usepackage[small]{caption}
\usepackage{times}
\usepackage{soul}
\usepackage{url}
\usepackage{graphicx}
\usepackage[table]{xcolor}
\usepackage{mathtools}
\usepackage{nicefrac}
\usepackage{amsfonts}
\usepackage{bm}
\usepackage{import}
\usepackage{booktabs}
\usepackage{algorithm}
\usepackage[noend]{algorithmic}
\usepackage{blindtext}
\usepackage{glossaries}
\usepackage{arydshln}
\usepackage[isbn=false,
 backend=biber,
 style=ieee,
 bibstyle=ieee, % write literature lists in IEEE style
 citestyle=numeric-comp, % \cite uses a numeric key
 sortcites=true, 
 maxbibnames=5,
 mincitenames=1,
 maxcitenames=2,
 giveninits=true, % use caps 
 backref=false]{biblatex}
\DeclareRobustCommand{\citet}[1]{\citeauthor{#1}~\cite{#1}}

\usepackage{siunitx}

\newtheorem{assumption}{Assumption}
\newtheorem{definition}{Definition}
\newtheorem*{remark}{Remark}

\newcommand{\fakepar}[1]{\vspace{1mm}\noindent\textbf{#1:}}

\newcommand*{\pdiff}[0]{\partial}

\newcommand{\eg}{e\/.\/g\/.,\/~}
\newcommand{\cf}{cf\/.\/~}

\newcommand{\fig}{Fig\/.\/~}
\newcommand{\defn}{Def\/.\/~}

\newcommand{\R}{\mathrm{I\!R}}
\newcommand{\argmin}{\arg\min}

\DeclareMathOperator{\cov}{\mathrm{cov}}

\makeatletter
\def\adl@drawiv#1#2#3{%
        \hskip.5\tabcolsep
        \xleaders#3{#2.5\@tempdimb #1{1}#2.5\@tempdimb}%
                #2\z@ plus1fil minus1fil\relax
        \hskip.5\tabcolsep}
\newcommand{\cdashlinelr}[1]{%
  \noalign{\vskip\aboverulesep
           \global\let\@dashdrawstore\adl@draw
           \global\let\adl@draw\adl@drawiv}
  \cdashline{#1}
  \noalign{\global\let\adl@draw\@dashdrawstore
           \vskip\belowrulesep}}
\makeatother

\newcommand{\cpara}{\bm{\theta}}

\usepackage{fancyhdr}
\newcommand{\mytitle}{To appear in the proceedings of the \textit{61st IEEE Conference on Decision and Control}.\\
\copyright 2022 IEEE. Personal use of this material is permitted. Permission from IEEE must be obtained for all other uses, in any current or future media, including reprinting/republishing this material for advertising or promotional purposes, creating new collective works, for resale or redistribution to servers or lists, or reuse of any copyrighted component of this work in other works.}

%% file: glossary.tex
\makeglossaries

\newacronym{tvbo}{TVBO}{time-varying Bayesian optimization}
\newacronym{gp}{GP}{Gaussian process}
\newacronym{bo}{BO}{Bayesian optimization}
\newacronym{b2p}{B2P}{Back-2-Prior}
\newacronym{ui}{UI}{Uncertainty-Injection}
\newacronym{lqr}{LQR}{linear quadratic regulator}
\newacronym{mab}{MAB}{multi-armed bandit}
\newacronym{vop}{VOP}{virtual observation point}
\newacronym{se}{SE}{squared-exponential}
\newacronym{mvn}{MVN}{multivariate normal distribution}
\newacronym{uitvbo}{UI-TVBO}{\gls{ui} in \gls{tvbo}}
\newacronym{ctvbo}{C-TVBO}{spatially convex \gls{tvbo}}
\newacronym{lcb}{LCB}{lower-confidence-bound}
\newacronym{nstvo}{NSTVO}{non-stationary time-varying optimization}
\newacronym{stvo}{STVO}{stationary time-varying optimization}

%% file: moin_matter/abstract.tex
\begin{abstract}

Changing conditions or environments can cause system dynamics to vary over time.
To ensure optimal control performance, controllers should adapt to these changes.
When the underlying cause and time of change is unknown, we need to rely on online data for this adaptation.
% A popular automatic controller tuning methods is \gls{bo} which directly models the control objective which sidesteps the need for online system identification.
In this paper, we will use \gls{tvbo} to tune controllers online in changing environments using appropriate prior knowledge on the control objective and its changes.
Two properties are characteristic of many online controller tuning problems:
First, they exhibit incremental and lasting changes in the objective due to changes to the system dynamics, \eg through wear and tear.
Second, the optimization problem is convex in the tuning parameters.
Current \gls{tvbo} methods do not explicitly account for these properties, resulting in poor tuning performance and many unstable controllers through over-exploration of the parameter space.
We propose a novel \gls{tvbo} forgetting strategy using \gls{ui}, which incorporates the assumption of incremental and lasting changes.
The control objective is modeled as a spatio-temporal \gls{gp} with \gls{ui} through a Wiener process in the temporal domain.
Further, we explicitly model the convexity assumptions in the spatial dimension through \gls{gp} models with linear inequality constraints.
In numerical experiments, we show that our model outperforms the state-of-the-art method in \gls{tvbo}, exhibiting reduced regret and fewer unstable parameter configurations.
\end{abstract}

\glsresetall

%% file: moin_matter/introduction.tex
\section{Introduction}

Automated controller tuning aims to find optimal controllers not only from model knowledge but also include data generated from the plant in the search.
An emerging method for controller tuning is \gls{bo} \cite{Shahriari_2016}.
It is well-suited for this task as it can incorporate a broad spectrum of prior knowledge about the control objective in a probabilistic model.
Most work on controller tuning with \gls{bo} has focused on tuning problems where the dynamics governing the system do not change over time \cite{Calandra_2016, Marco_2016,Neumann_2020,berkenkamp2021bayesian, muller2021local, Doerschel_2021, brunke2022safe}.
However, in practice, control performance might decay when the underlying system changes, \eg due to wear and tear or payload differences.
Herein we consider changes that are unknown a priori so that tuning must be online and therefore sample efficient.
\begin{figure}[t]
    \centering
    \includegraphics[]{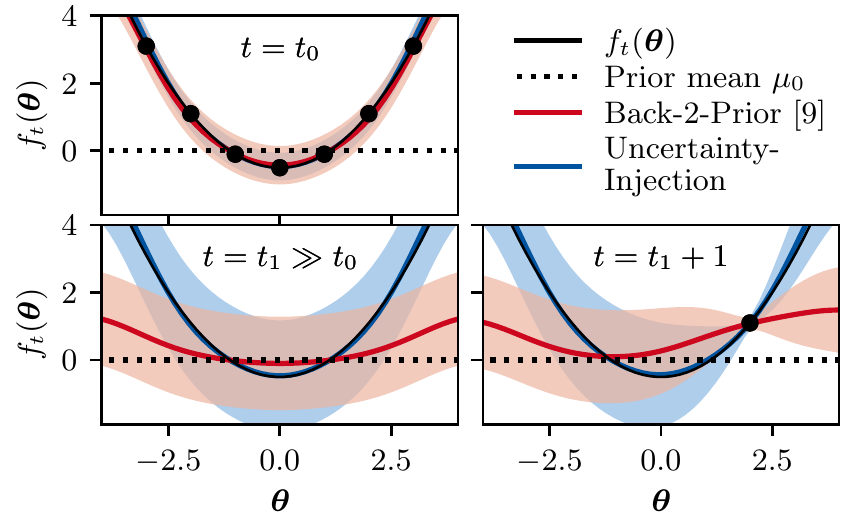}
  \caption{Intuition behind the proposed forgetting strategy (blue) in comparison to the state-of-the-art in \acrshort{tvbo} (red). At $t_0$, recent data of the control objective is available (top).
  Over time forgetting occurs ($t_1$, bottom left) and uncertainty increases if no new data is added.
  Existing approaches assume a stationary spatio-temporal process whose posterior falls \textit{back to the prior} (B2P) over time (red). In contrast, we propose modeling the control objective as a non-stationary spatio-temporal process via \textit{uncertainty injection} (UI), better capturing engineering knowledge and intuition. The model can remember relevant past information in the form of the posterior mean (blue). As new data is added ($t_1+1$, bottom right), this information can be leveraged by a \acrshort{tvbo} algorithm as it does not need to re-explore the whole domain.}
  \label{fig:forgetting_strategies}
%   \vspace{-0.3cm}
\end{figure}

In general, the time-varying optimization problem with arbitrary temporal changes is infeasible, and we need some regularity assumptions to derive meaningful algorithms.
In \gls{tvbo} \cite{Bogunovic2016} these are formulated as a prior distribution that defines correlations of the objective over time.
This prior knowledge restricts the temporal change of the objective function and thus guides the optimization algorithm in the exploration-exploitation trade-off when learning about changes.
Especially in online controller tuning, we want to include as much prior information as possible to avoid over-exploration and increase sample efficiency, performance, and safety.

In this work, we distinguish between two different kinds of time variations: variations in the objective function (modeled as a sample from a stochastic process) and variations of the process itself. For clarity, we reserve the word \emph{stationary process} to qualify the second kind of time variations. Therefore, a time-varying objective function can either result from a stationary or a non-stationary process (see \defn\ref{def:non-stationary}).
Current \gls{tvbo} methods exclusively focus on stationary processes, which exhibit a catastrophic forgetting behavior over time as the posterior prediction falls back to the prior, resulting in excessive exploration when tuning controllers.
In the controller tuning setting, this \gls{b2p} forgetting assumes that changes in the system dynamics always revert to their prior expectation and thus are mean-reverting.
However, temporal changes in control are usually caused by incremental and lasting changes such as wear and tear.
For example, friction only tends to increase over time and is usually not a mean-reverting process.
In addition, mean-reversion leads to undesirable exploration behavior when combined with a constant prior mean, another prevalent assumption in \gls{bo}.
In that case, \gls{tvbo} has to explore the whole domain over and over again.
Instead of \gls{b2p} forgetting, we propose a novel \gls{tvbo} forgetting strategy using \gls{ui}, which incorporates the assumption about incremental and lasting changes in the control objective.
Specifically, \gls{ui} results in a non-mean reverting process which is beneficial, for example, when the dynamics and control performance changes over time due to increased friction.
We also directly incorporate prior knowledge of the objective's convexity as shape constraints for efficient exploration to further guide the optimization.

\fakepar{Contribution}
In summary, we propose a novel model for \gls{tvbo} that encodes relevant assumptions for online controller tuning, a non-stationary time-varying optimization problem. In particular, we propose to use \gls{ui} instead of \gls{b2p} (see \fig\ref{fig:forgetting_strategies}).
With \gls{ui} in \gls{tvbo}, the model remembers the posterior mean over time while allowing for a wide variety of changes in the objective, such as drifts and jumps, without over-exploration.
This better captures our prior knowledge about temporal changes in control systems.
Additionally, accounting for many convex control objectives, we explicitly incorporate a priori knowledge about convexity into the \gls{tvbo} algorithm.
The convexity assumption allows global extrapolation from local queries.
The main contributions are:
\begin{itemize}
    \item \emph{\gls{uitvbo}}: A surrogate model for \gls{tvbo} modeling non-stationary temporal change in the objective functions using a Wiener process.
    \item We use linearly inequality constrained \glspl{gp} \cite{Agrell_2019} to model convex time-varying objective functions for \gls{tvbo}.
\end{itemize}

We show, based on an \gls{lqr} problem under non-stationary changes, that \gls{uitvbo} needs to explore less and, thereby, yields improved regret compared to the state-of-the-art.
Additionally, including the prior knowledge of convexity further reduces the regret and prevents unstable parameter configurations.

\section{Problem Formulation}\label{sec:problem}
We seek to sequentially optimize controller parameters $\cpara_t$ of an unknown time-varying objective function $f_t\colon \mathcal{X}\times\mathcal{T} \mapsto \R$.
%with $f_t(\cpara) \coloneqq f(\cpara, t)$. 
Thus, the optimization problem can be written as
\begin{equation}
    \cpara_t^* = \argmin_{\cpara \in \mathcal{X}} f_t(\cpara)
    \label{eq:tvbo_setup}
\end{equation}
at the discrete time step $t\in \mathcal{T} = \{1,2, \dots\}$ within a feasible set $\cpara \in \mathcal{X} \subset \R^D$. At each time step, an algorithm can query the objective function based on past information at a chosen location $\hat\cpara_t$ and obtains a noisy observation of the form
$
  y_t = f_t(\cpara) + w_t
$
with zero mean i.i.d. Gaussian noise $w_t \sim \mathcal{N}\left(0, \sigma_n^2 \right)$ where $\sigma_n^2$ is the noise variance.
The performance of the tuning algorithm is defined in terms of regret measuring the difference between optimal and chosen parameters as formalized in the following definition.
\begin{definition}[Regret (\cf\cite{Bogunovic2016})]\label{def:regret}
Let $\cpara_t^*$ be the optimum of the time-varying function $f_t(\cpara)$ as $\cpara_t^* = \argmin_{\cpara \in \mathcal{X}} f_t(\cpara)$ at time step $t$ and suppose $\hat{\cpara}_t$ is queried at time step $t$. Then the regret after $T$ time steps is $R_T \coloneqq \sum_{t=1}^T \left(f_t(\hat{\cpara}_t) - f_t(\cpara_t^*)\right)$.
\end{definition}

Since solving \eqref{eq:tvbo_setup} for arbitrary changes in $f_t$ is not feasible, we must formulate some assumptions.
We assume that the objective function can be modeled as a \gls{gp} \cite{Rasmussen_2006}.
Where applicable, we assume spatial convexity of the objective function modeled as an inequality constraint \gls{gp} \cite{Agrell_2019}.
\begin{assumption}[Smoothness and temporal change]\label{ass:gp}
The objective function $f_t$ is a sample path from a known GP prior $\phi \sim \mathcal{GP}(\mu_0, k)$ with a separable spatio-temporal kernel $k\colon \mathcal{X} \times \mathcal{T} \mapsto \R$ as
\begin{equation}
    k((\cpara,t),(\cpara',t')) = k_{S}(\cpara, \cpara') k_{T}(t, t'). \label{eq:spatiotemporal}
\end{equation}
Moreover, the mean function $\mu_0$ is at least twice differentiable, and the covariance function $k$ is at least four times differentiable, both w.r.t. $\cpara$.
\end{assumption}

\begin{assumption}[Convexity]\label{ass:prior_knowledge_convex}
Every realized sample $f_t$ from $\phi$ is at least twice differentiable with respect to $\cpara$ and the Hessian $\nabla^2_{\cpara} f_t$ is semi positive-definite $\forall t \in \mathcal{T}$ and $\forall \cpara \in \mathcal{X}$. 
\end{assumption}
Many relevant controller tuning problems can be formulated as convex optimization problems (\eg the \acrshort{lqr} problem) making Assumption~\ref{ass:prior_knowledge_convex} especially suitable in the context of control systems.
For further clarity, we distinguish between two types of time-varying optimization problems, depending on the stationarity of the process.
\begin{definition}\label{def:stationary_process}
For a stationary spatio-temporal process it holds that for all $\cpara \in \mathcal{X}$, $\tau, t_1,\dots,t_n \in \mathbb{R}$, and $n\in \mathbb{N}$
$
   P(\phi_{t_1}(\cpara), \dots, \phi_{t_n}(\cpara)) = P(\phi_{t_1+\tau}(\cpara), \dots, \phi_{t_n+\tau}(\cpara)). 
$\label{eq:stationary}
\end{definition}
\begin{definition}\label{def:stationary}
\Gls{stvo} problems are optimization problems of the form \eqref{eq:tvbo_setup} under Assumption~\ref{ass:gp} where $\phi$ is a stationary spatio-temporal process as in \defn\ref{def:stationary_process}.
\end{definition}

\begin{definition}\label{def:non-stationary}
\Gls{nstvo} problems are optimization problems of the form \eqref{eq:tvbo_setup} under Assumption~\ref{ass:gp} where $\phi$ is a non-stationary spatio-temporal process.
\end{definition}

To the best of our knowledge, prior work in \gls{tvbo} is based on \cite{Bogunovic2016} and, therefore, considers problems of the form in Def.~\ref{def:stationary}.
Instead, the approach presented herein explicitly models \gls{nstvo} problems (Def.~\ref{def:non-stationary}), as these often occur in physical systems (\eg due to wear and tear or load changes).

%% file: moin_matter/related_work.tex
\section{Related Work}

We propose a spatio-temporal model for online controller tuning using \gls{tvbo} and utilize prior knowledge about the shape of the objective function to guide the optimization. 
We discuss prior work on \gls{tvbo}, \gls{bo} for controller tuning, and shape constraints on the surrogate model in \gls{bo}.
We also briefly touch on probabilistic models for time-varying functions.
We omit work on time-varying convex optimization that is not in the \gls{bo} setting and refer the reader to \citet{Simonetto_2020} for a recent overview.

\fakepar{Time-Varying Bandit Optimization}
Minimizing regret over finite actions in a time-varying environment is studied in dynamic \glspl{mab}.
Here, both forgetting strategies -- UI \cite{Slivkins_2008,Besbes_2014} as well as \gls{b2p} \cite{Chen_2021} -- have been considered.
GP-based \gls{tvbo} with a spatio-temporal process can be considered as a special case of contextual \gls{bo} \cite{Krause_2011} but it was first explicitly discussed by \citet{Bogunovic2016}.
The authors introduce a stationary spatio-temporal process as well as a time-varying UCB variant with regret bounds to solve \gls{stvo} problems.
The time-varying UCB variant increases the UCB exploration parameter over time, which leads to undesirable high exploration in controller tuning problems and is difficult to tune in practice.
The work of \citet{Bogunovic2016} has been adapted to different applications in low-dimensional settings with up to four parameters, such as controller learning \cite{Su_2018}, safe adaptive control \cite{Koenig_2021}, and online hyperparameter optimization \cite{Parker-Holder_2020}.
In contrast to the algorithms based on \citet{Bogunovic2016}, we use the standard UCB algorithm for \gls{tvbo} without increasing exploration over time.
The main difference between \gls{uitvbo} and the methods based on \citet{Bogunovic2016} is that \gls{uitvbo} intends to solve \gls{nstvo} problems such as controller tuning.
%We present the first approach in \gls{tvbo} to model the temporal dimension using the concept of \gls{ui}.

\fakepar{\gls{bo} for Controller Tuning}
\gls{bo} has emerged as a powerful and popular method to tune controllers given only limited system knowledge (\eg \cite{Calandra_2016, Marco_2016,Neumann_2020,berkenkamp2021bayesian, muller2021local, Doerschel_2021, brunke2022safe}).
However, controller tuning with changing dynamics is significantly less explored in literature. 
\citet{Koenig_2021} considers an adaptive control problem for safe model-free adaptive control where they adopt the setting of \citet{Bogunovic2016}.

\fakepar{Shape Constraints in \gls{bo}}
The main benefit of incorporating prior knowledge in \gls{bo} is an increase in sample efficiency as it reduces the hypothesis space of the objective function. 
A natural prior in engineering applications are shape constraints which have been used in \gls{bo} surrogate models by \citet{Jauch_2016}. 
They employ shape-constrained models based on \glspl{gp} introduced by \citet{Wang_2016} and present experimental results for up to two-dimensional optimization problems.
We use the framework for GPs under linear inequality constraints more recently proposed by \citet{Agrell_2019}.
However, due to the need to sample from high-dimensional truncated normal distributions, these models are limited to low-dimensional parameter spaces and experimental results therein are at most two-dimensional.
To the best of our knowledge, we are the first to incorporate shape constraints in \gls{tvbo}.
An alternative approach is presented by \citet{Marco_2017} in which they design a kernel function for the convex \gls{lqr} tuning problem that incorporates uncertainty about the linear model.
However, it is only applicable to systems with one-dimensional states and inputs.

\fakepar{Probabilistic Models for Time-Varying Functions}
Modeling temporal and spatio-temporal processes with a \gls{gp} is a topic of interest in regression and Bayesian filtering \cite{Vaerenbergh_2012,Sarkka_2013,Carron_2016}. 
Analogous to their approaches, we utilize a separable spatio-temporal kernel.
\citet{Vaerenbergh_2012} further discussed the remembering-forgetting trade-off when modeling a time-varying function and introduced the terms \gls{b2p} and \gls{ui}.
While \gls{b2p} has been the implicit forgetting strategy in prior work on \gls{tvbo}, we argue that \gls{ui} is a more suitable modeling approach for many practical problems such as online controller tuning.

%% file: moin_matter/methods.tex
\section{Methods}

\begin{algorithm}[b]
\caption{TVBO algorithm.}
\begin{algorithmic}[1]
\STATE \textbf{Input:} $\mathcal{GP}(0, k)$, $\mathcal{X} \subset {\rm I\!R}^D$, $\mathcal{D}_N=\left\{ (t_j,\cpara_j,y_j)\right\}_{j=0}^N$
\FOR{$t = N+1, N+2, \dots$ }
\STATE Train GP model with $\mathcal{D}_t$
\STATE Optimize $\hat\cpara_{t+1} = \argmin_{\cpara\in \tilde{\mathcal{X}}_{t+1}} \alpha(\cpara, t+1|\mathcal{D})$
\STATE Query objective function $y_{t+1} = f_{t+1}(\hat\cpara_{t+1}) + w$
\STATE $\mathcal{D}_{t+1} = \mathcal{D}_{t} \cup \{(y_{t+1}, \hat\cpara_{t+1}, t+1)\}$
\ENDFOR
\end{algorithmic}
\label{alg:algorithm}
\end{algorithm}

Motivated by the online controller tuning problem, we present a non-stationary surrogate model for \gls{tvbo} based on \gls{ui}, \gls{uitvbo}.
Furthermore, we present an approach to incorporate Assumption~\ref{ass:prior_knowledge_convex} into \gls{tvbo}.
The \gls{tvbo} algorithm is shown in Algorithm~\ref{alg:algorithm}.

\subsection{Uncertainty Injection in TVBO}\label{sec:ui}
In contrast to \gls{b2p} forgetting, \gls{ui} simply adds uncertainty about the objective function over time.
\gls{ui} can be implemented by modeling $f_t$ with a process that is not mean-reverting and whose variance increases with time when no data is received.
We choose to use a Wiener process, in which the expected value remains the same over time while the variance increases linearly.
The Wiener process kernel is
\begin{equation}
    k_{T,wp}(t,t') = \sigma_w^2 \left( \min(t,t') - c_0\right)
    \label{eq:wienerprocesskernel}
\end{equation}
where $c_0$ and $\sigma_w^2$ are hyperparameters that model the magnitude of change in the temporal dimension.
Following Assumption~\ref{ass:gp}, we use a spatio-temporal product kernel $k$ for \gls{uitvbo}.
With the \gls{se} kernel as $k_S$, this results in
\begin{align}
    k((\cpara,&t),(\cpara',t')) = k_{S,se}(\cpara,\cpara') \cdot k_{T,wp}(t,t') \nonumber\\
    &= \sigma_k^2\exp\left[-\frac{1}{2} \bm{r}^T \boldsymbol{\Lambda}^{-1} \bm{r}\right] \cdot \sigma_w^2 \left( \min(t,t') - c_0\right)
    \label{eq:ui-tvbo_model}
\end{align}
where $\bm{r} = \cpara - \cpara'$, $\boldsymbol{\Lambda} = \mathrm{diag}(\sigma_{l,1}^2, \dots, \sigma_{l,D}^2)$ are the lengthscales $\sigma_{l,i}^2$, and $\sigma_k^2$ is the output variance.
By definition, at time step $t=0$, no forgetting has occurred.
Therefore, the output of the spatio-temporal kernel at $t=0$ must be the output of the spatial kernel. Thus, for any $t' >0$
\begin{equation}
    k_{T,wp}(0,t') = \sigma_w^2 \left(\min(0,t') - c_0\right) = 1
\end{equation}
should hold, which implies that $c_0 = - \sigma_w^{-2}$.
At different time steps $t'$ with $0<t<t'$, assuming $t$ is the time step of the last query, the variance should increase linearly in time based on a forgetting factor $\hat{\sigma}_w^2$ as
$\hat{\sigma}_w^2 (t'-t)$ to be consistent with the model of a Wiener process.
The posterior covariance for $r = 0$, where $k_{S,se} = \sigma_k^2$, is given by
\begin{align}
    \cov[(\cpara,t),(\cpara',t')]
    =& \sigma_k^2 k_{T,wp}(t',t') - \sigma_k^2 k_{T,wp}(t',t) \nonumber\\
    &\left(\sigma_k^2 k_{T,wp}(t,t)\right)^{-1} \sigma_k^2 k_{T,wp}(t,t') \nonumber\\
    =&\sigma_k^2 \sigma_w^2\left(t' - t\right). \label{eq:var}
\end{align}
We thus re-parametrize the increase in variance as $\hat{\sigma}_w^2$ with $\hat{\sigma}_w^2 = \sigma_k^2\sigma_w^2 $ for \gls{uitvbo}.
Note, that, as desired, for $\sigma_w^2 \to 0$ \gls{uitvbo} converges to the time-invariant model as $\lim_{\sigma_w^2 \to 0} k_{T,wp} = 1$ with the proposed parametrization.
We formalize the assumptions about the temporal changes when using a Wiener process in Assumption~\ref{ass:wiener}.
\begin{assumption}[Temporal Wiener process]\label{ass:wiener}
$\phi_t(\cpara)$ follows a Wiener process with known variance $\hat\sigma_w^2$ at all $\cpara$.
Well known results mean that $\delta(\cpara) = \phi_{t+1}(\cpara) - \phi_{t}(\cpara)$ at point $\cpara$ is an i.i.d. normal distribution $\delta(\cpara) \sim \mathcal{N}(0,\hat\sigma_w^2)$.
\end{assumption}

%%%%%%%%%%%%%%%%%%%%%%%%%%%%%%%%%%%%%%%%%%%%%%%%%
\subsection{Modeling Convex Time-Varying Objective Functions}\label{sec:convex}

In online controller tuning, it is desirable to limit the exploratory behavior of the algorithm as it directly influences the performance. This can be achieved by embedding prior knowledge about the shape of the objective function into \gls{tvbo}. Since many control objectives are convex, we formulate a surrogate model that approximates Assumption~\ref{ass:prior_knowledge_convex}.
When choosing the controller parameters for $t+1$, we have a current estimate of the optimum $\hat\cpara^*_t$ and want to find a good $\hat\cpara^*_{t+1}$.
For this, we optimize the \gls{bo} acquisition function operating on the posterior of $\phi_{t+1}(\cpara)$.
Since, by Assumption~\ref{ass:prior_knowledge_convex}, the objective is convex, we modify the posterior using inequality constraints on the second derivative for each dimension using the method proposed by \citet{Agrell_2019}.
For more than one dimension, this is only an approximation of Assumption~\ref{ass:prior_knowledge_convex} as it enforces convexity only in coordinate directions but does not enforce the Hessian to be positive semi-definite.

The inequality constraints are enforced on a finite set of \glspl{vop}.
Given a sufficient amount of \glspl{vop} combined with the smooth \acrshort{se} kernel, we can approximate convexity in the feasible set $\mathcal{X}$.
In order to sample from the constrained posterior (\cf \cite[Algorithm~3]{Agrell_2019}), we use the sampling method introduced by \citet{Botev2016} but switch for computational reasons to a Gibbs sampling algorithm \cite{Wang_2016} if the distribution is high-dimensional or the covariance matrix close to singular.

Since the acquisition function is optimized only at time step $t+1$ of \gls{tvbo}, it is sufficient to place \glspl{vop} in $\mathcal{X}$ only at $t+1$.
Ideally, the \glspl{vop} are placed densely within $\mathcal{X}$.
However, sampling becomes computationally expensive for large numbers of \glspl{vop} \cite{Agrell_2019}.
This limits constraining the posterior to only low-dimensional problems.
However, it is still suitable for PID-controller and low-dimensional \gls{lqr} tuning. We will consider the latter in the numerical results.

\begin{figure}[t]
  \centering
  \includegraphics[]{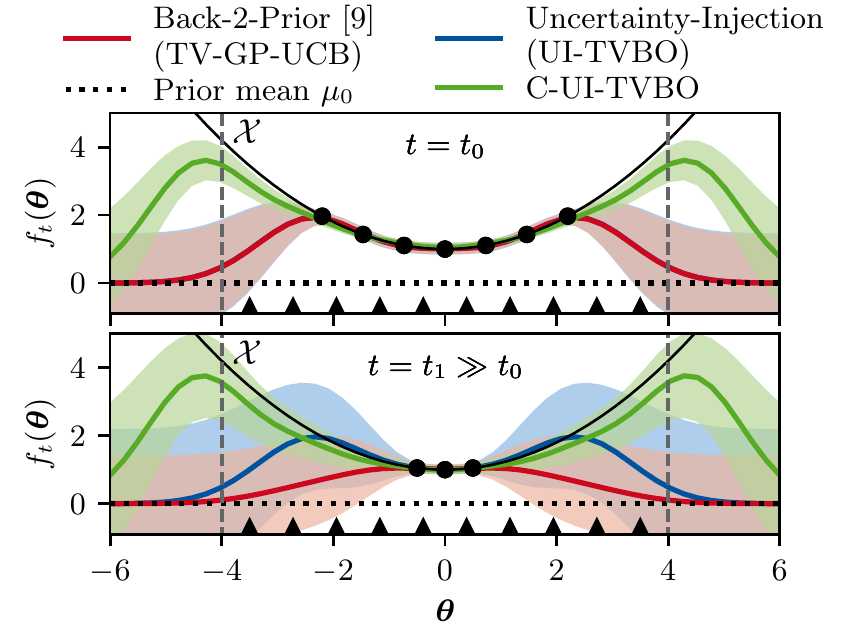}
  \caption{Visualization of the constrained (green) and unconstrained posterior distributions (red and blue) with the same \acrshort{se} kernel. The triangles denote the \glspl{vop} at which the convexity is enforced. As in \fig\ref{fig:forgetting_strategies}, \gls{uitvbo} maintains past information.
  Further adding the constraints (C-\gls{uitvbo}) allows the model to extrapolate within $\mathcal{X}$ (gray dashed lines) from data near the optimum while away from the \glspl{vop} the constrained posterior converges back to the unconstrained posterior.
  Both \gls{ui} and the constraints allow for increased uncertainty but retain critical \textit{structural} information about the optimization and guide the exploration of \gls{tvbo}.}
  \label{fig:constrained_GP}
  \vspace{-0.3cm}
\end{figure}

The method of \citet{Agrell_2019} requires the derivatives of the kernel w.r.t. to $\cpara$. We obtain them by leveraging the separable spatio-temporal kernel in \eqref{eq:spatiotemporal}. Since the temporal kernel is independent of $\cpara$, we can write ${\frac{\pdiff}{\pdiff \cpara}k = (\frac{\pdiff}{\pdiff \cpara} k_S) k_T}$. Calculating the necessary Gram matrices is then straightforward \cite[Section~4.2]{Agrell_2019}.
This concept of constraining the posterior in a time-varying context is visualized in \fig\ref{fig:constrained_GP}.
For large domains, we avoid computational bottlenecks by enforcing local convexity. 
We place equidistant \glspl{vop} in a subset of the feasible set $\tilde{\mathcal{X}}_{t+1} \subseteq \mathcal{X}$ around the current best estimate $\hat{\cpara}^*_t$. The acquisition function is then optimized in $\tilde{\mathcal{X}}_{t+1}$.
\begin{remark}
If $\cpara_t^*$ is in $\tilde{\mathcal{X}}_t$, the regret is unaffected by the local optimization. 
This is usually true for gradually changing processes.
Nevertheless, depending on the rate of change $\hat{\sigma}^2_w$, this might not hold.
In our empirical evaluation, we observed that even if the optimizer jumps out of $\tilde{\mathcal{X}}_t$, the algorithm will recover and find a $\tilde{\mathcal{X}}_t$ that contains the optimum.
\end{remark}

The method presented in this section is independent of the temporal process of $\phi$ and can be used for all modeling approaches in \gls{tvbo}.
Hence, we will compare constrained variations for both \gls{b2p} (C-TV-GP-UCB) and \gls{ui} (C-\gls{uitvbo}) in the numerical results.

%% file: moin_matter/results.tex
\begin{figure*}[t]
    \centering
    \includegraphics[]{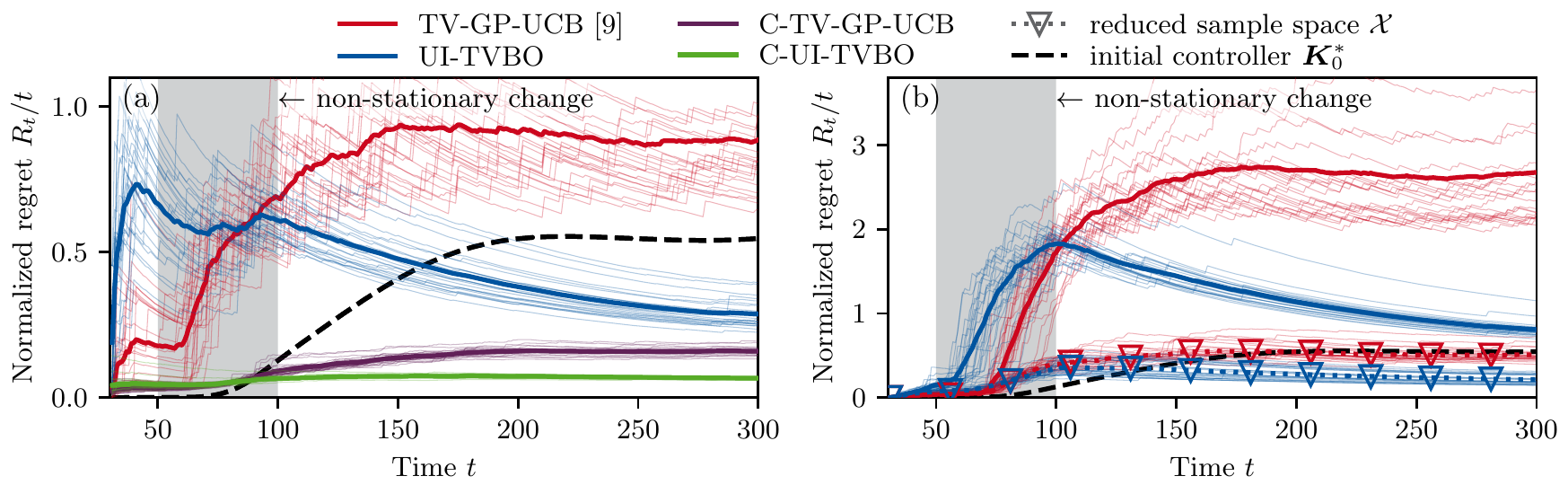}
    \caption{Normalized regret of the (a) 2D \gls{lqr} problem and (b) 4D \gls{lqr} problem.
    The thin lines denote the $25$ individual runs and the thick line is the mean. 
    After the change in friction according to \eqref{eq:friction} (gray area), the regret of TV-GP-UCB increases both in (a) and (b) due to the \gls{b2p} stationarity assumption on the temporal process of $\phi$.
    In contrast, \gls{uitvbo} displays asymptotically decreasing regret outperforming TV-GP-UCB. If the bounds of $\mathcal{X}$ are reduced to consider only stable parameter configurations in (b) (blue with triangles), \gls{uitvbo} also outperforms $\bm{K}_0^*$.
    Including the constraints in (a) further decreases the regret and its variance substantially.}
    \label{fig:Regret4D_over_time}
    \vspace{-0.4cm}
\end{figure*}

\section{Numerical Results}

To evaluate our modeling approach, we benchmark it against the state-of-the-art model TV-GP-UCB with a forgetting factor of $\varepsilon \in (0,1)$ \cite{Bogunovic2016} on a \gls{lqr} tuning problem.
The \gls{lqr} problem is a classic problem in optimal control theory controlling a linear dynamical system by minimizing a quadratic cost function.
It satisfies Assumption~\ref{ass:prior_knowledge_convex} making it suitable to test our methods.
In addition to TV-GP-UCB, we compare our method against the initially optimal state-feedback control as a baseline to investigate whether tuning actually improves performance.

For the system at hand, we consider a cart-pole system linearized around the upper equilibrium point.
%with the goal of balancing the pole in the upright position.
The linear state space model is the following:
\begin{equation}
    \bm{\dot{x}} =\begin{bmatrix}
         0 & 1 & 0 & 0 \\
         0 & -\frac{1}{T_1} & 0 & 0\\
         0 & 0 & 0 & 1\\
         0 & \frac{1}{2}\frac{m_p l}{J_d T_1} &
     \frac{1}{2} \frac{m_p l g}{J_d} & -\frac{\tau_p(t)}{J_d}
    \end{bmatrix}\bm{x} + \begin{bmatrix}
    0 \\
    \frac{K_u}{T_1} \\
    0 \\
    -\frac{1}{2}\frac{m_p l}{J_d}\frac{K_u}{T_1}
    \end{bmatrix} u
\end{equation}
with $\bm{x} = [x, \dot{x}, \varphi, \dot{\varphi}]$, where $x$ is the cart position and $\varphi$ the pole angle. Furthermore, ${m_p=0.0804\si{\kilogram}}$, $l=0.147\si{\meter}$, $J_d=0.5813\cdot 10^{-3}\si{\kilogram\meter}^2$, $\tau_p(t)$ are the mass, length, moment of inertia, and time varying-friction of the pole, respectively.
The input $u$ is the desired cart velocity and we choose ${T_1=1\si{\second}}$ and $K_u=1$.
To induce a non-stationary smooth change in the dynamics model, the initial friction $\tau_{p}^0 \coloneqq \tau_{p}(0)=2.2\cdot10^{-3}\si{\newton\meter\second}$ in the bearing is increased at $t_1 = 50$ until $t_2 = 100$ as 
\begin{equation}
    \tau_p(t) = \begin{cases}
        \tau_{p}^0 ,& t < t_1 \\
        \tau_{p}^0+\frac{3}{2} \tau_{p}^0 \left(1-\cos\left(\frac{\pi}{t_1}(t-t_1)\right) \right),& t_1 \leq t \leq t_2 \\
        3 \tau_{p}^0 + \frac{1}{2}\tau_{p}^0 \sin\left(-\frac{\pi}{t_2}t\right) ,& t > t_2 \\
        \end{cases} \label{eq:friction}
\end{equation}
A lasting increase in friction is a realistic real-world scenario caused by, \eg wear and tear.

We choose the objective function to be the approximate \gls{lqr} cost over a simulation horizon $M$ as
\begin{align}
    \hat J_t =\frac{1}{M} \left[ \sum_{{m}=0}^{M-1} \bm{x}_{m}^T \bm{Q}\bm{x}_{m} +  u_{m}^T R u_{m} \right] \label{eq:lqr_cost}
\end{align}
with the positive definite weight matrices $\bm{Q}\in\R^{4\times4}$ and ${R\in\R}$, and where $m$ is a discrete simulation time step (\textit{not} a \gls{tvbo} time step). Due to the change in friction over time, the cost function in \eqref{eq:lqr_cost} is time-varying.

The state-feedback controller at time step $t$ is defined as $u_m = - \bm{K}_t^T \bm{x}_m$.
We directly optimize the controller gains $\theta_t^i$ as $\cpara_t \coloneqq \bm{K}_t = [\theta_t^0,\theta_t^1,\theta_t^2,\theta_t^3]$.
At each time step $t$ of \gls{tvbo}, we perform a closed-loop simulation given the chosen parameters of the controller over $M=1000$ discrete time steps with a sampling time of $0.02\si{\second}$. The accumulated cost is then calculated using \eqref{eq:lqr_cost} with $\bm{Q}=10\cdot \bm{I}$ and ${R=1}$, and the result is provided as noisy observation (due to process noise) to the \gls{tvbo} algorithm.
% Since we choose $\bm{Q}$ to be diagonal, enforcing convexity only in coordinate direction as stated in Sec.~\ref{sec:convex} is sufficient.
Given the linear state space model, it is possible to calculate the optimal state-feedback controller at each time step $\bm{K}^*_t$ using the algebraic Riccati equation if the system dynamics were known. %but we do not assume any knowledge about the system dynamics beyond the collected cost.

We use the standard \gls{tvbo} algorithm shown in Algorithm~\ref{alg:algorithm} and choose LCB as the acquisition function as
% \begin{equation}
%   {\alpha(\cpara, t+1|\mathcal{D}) = \mu_{t+1}(\cpara) - \sqrt{\beta_{t+1}}\, \sigma_{t+1}(\cpara)}  
% \end{equation}
$
  {\alpha(\cpara, t+1|\mathcal{D}) = \mu_{t+1}(\cpara) - \sqrt{\beta_{t+1}}\, \sigma_{t+1}(\cpara)}  
$
with an exploration-exploitation factor of $\beta_{t+1}=2$. By setting $\beta_{t+1}$ to a constant, we omit its poly-logarithmic increase as in \citet{Bogunovic2016}. Especially in controller tuning, it is undesirable to increase explorative behavior over time due to the increased risk of the controller becoming unstable. For convex function it is also unnecessary. We further set the forgetting factors to $\hat{\sigma}_w^2=\varepsilon=0.03$, which showed good empirical performance for all approaches in preliminary experiments.
The algorithm is implemented using BoTorch~\cite{Balandat_2020} and we use GPyTorch~\cite{Gardner_2018} for the \gls{gp} surrogate model\footnote{Code and parameters published at {\url{https://github.com/brunzema/uitvbo}}.}.

We consider both a 2D and a 4D \gls{lqr} problem. For the 2D problem, we set $\theta_t^0$ and $\theta_t^1$ to their optimal values (i.e., the first two entries of $\bm{K}_t^*$).
At the start, we are given an initial set of $N$ parameters $\mathcal{D}_N=\{ (t_j,\hat\cpara_j,y_j)\}_{j=0}^N$ where $y_j$ and $t_j$ are the corresponding observations and time stamps.
For both the 2D \gls{lqr} problem and 4D \gls{lqr} problem, the initial set is composed of $N=30$ stable controller parametrizations, and the performance is normalized to a mean of zero and unit variance.
The feasible set $\mathcal{X}$ is chosen as a hyperbox, also allowing unstable controller configurations.
We determine unstable controllers heuristically via a cost threshold.
In order to not deteriorate the hyperparameter optimization of the \gls{gp}, the observations for unstable controller configurations $\bar{\cpara}_t$ are based on the current posterior as $y_t(\bar{\cpara}_t) = \mu_t(\bar{\cpara}_t) + 3\sigma_t(\bar{\cpara}_t)$.
Intuitively, this means the cost for an unstable controller is set ``as high as possible'' given the likelihood function of the current posterior.
To guide the hyperparameter optimization, we use Gamma priors on the length scales of the spatial kernel.

Table~\ref{tab:results_summary} shows the results of $25$ different runs of both problems each with a time horizon of $T=300$.
\begin{table}[b!]
\caption{Summary of the results of $25$ runs with $T=300$.}
\label{tab:results_summary}
\begin{center}
\begin{tabular}{c c c}
\toprule
& & \textbf{No. of unstable}\\
\textbf{Variation} &  \textbf{Regret} \textbf{($\mu \pm \sigma$)}&  \textbf{controller }\textbf{($\mu \pm \sigma$)}\\
\midrule
Baseline $\bm{K}_0^*$   & $164.09$  & -- \\
\midrule
\midrule
\multicolumn{3}{c}{\textit{2D \gls{lqr} problem}}\\
\midrule
TV-GP-UCB   & $265.94 \pm 71.51$  & $11.72 \pm 1.90$ \\
UI-TVBO   & $86.1 \pm 13.48$  &$2.0 \pm 0.86$ \\
\cdashlinelr{1-3}
C-TV-GP-UCB   & $47.85 \pm 3.96$  &$\bm{0} \pm \bm{0}$ \\
\textbf{C-UI-TVBO}   & $\bm{19.97} \pm \bm{1.17}$  &$\bm{0} \pm \bm{0}$ \\
\midrule
\multicolumn{3}{c}{4D \textit{\gls{lqr} problem}}\\
\midrule
TV-GP-UCB   & $802.45 \pm 241.41$ & $17.84 \pm 3.87$\\
\textbf{UI-TVBO}   & $\bm{242.40} \pm \bm{26.19}$  &$\bm{4.28} \pm \bm{1.70}$\\
\midrule
\multicolumn{3}{c}{\textit{4D \gls{lqr} problem (reduced sampling space)}}\\
\midrule
TV-GP-UCB   & $151.78 \pm 21.68$  & by design $0$\\
\textbf{UI-TVBO}   & $\bm{64.17} \pm \bm{15.79}$  & by design $0$\\
\bottomrule
\end{tabular}
\end{center}
\end{table}
For calculating the regret (\defn\ref{def:regret}), unstable iterations are skipped since $f_t(\hat\cpara_t) - f_t(\cpara_t^*)$ is not well defined for the corresponding iteration. Note that this favors the approach with more unstable controllers, which in all cases is TV-GP-UCB.
In both the 2D and 4D \gls{lqr} problem, our approach \gls{uitvbo} significantly outperforms the state-of-the-art in terms of regret, yields a smaller variance and drastically reduces the number of unstable controllers. 
Furthermore, \gls{uitvbo} shows smaller regret than the initial optimal controller $\bm{K}_0^*$ in the 2D problem, whereas TV-GP-UCB does not.
Due to the large sampling space, both methods cannot outperform $\bm{K}_0^*$ in the 4D problem.
Consequently, no adaptive tuning would have resulted in a lower overall cost.
If we tighten the bounds on the feasible set to consider only stable parameter configurations (bottom, Table~\ref{tab:results_summary}), \gls{uitvbo} again outperforms $\bm{K}_0^*$ in 4D.
Including the convexity constraints prevents unstable configurations and further decreases the regret and its variance.
This suggests that prior knowledge about the convexity of the problem can be enough to make controller tuning safe in some instances.

To further compare the behavior of the models, \fig\ref{fig:Regret4D_over_time} shows the normalized mean regret (thick lines) as well as the individual runs (thin lines) over time. Following the non-stationary change, TV-GP-UCB further increases in normalized regret and individual runs display a sawtooth-like behavior caused by excessive exploration of \gls{b2p} forgetting.
In contrast, \gls{uitvbo} (empirically) displays the desirable asymptotically decreasing regret \cite{Bogunovic2016, Krause_2011}.

\begin{figure}[t]
  \centering
  \includegraphics[]{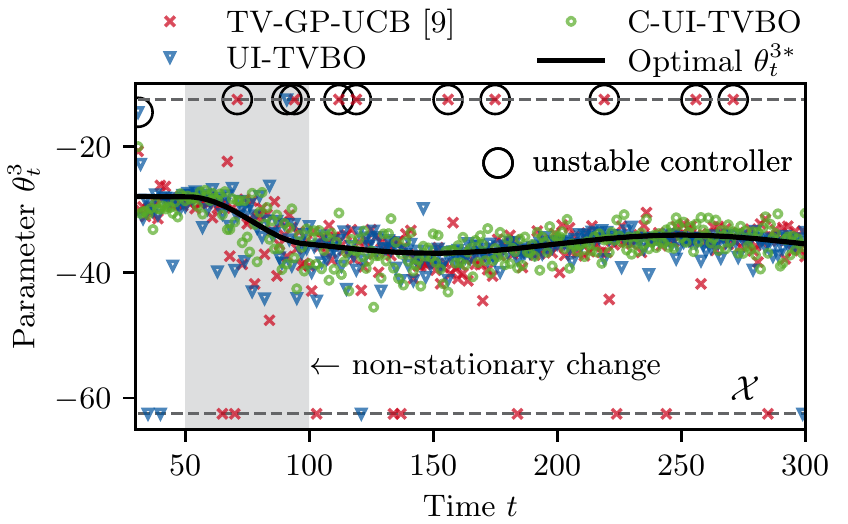}
  \caption{Query trajectories of the different variations (2D \gls{lqr} problem). Note that every variation chooses the same amount of queries, but some markers might be covered. After the non-stationary change resulting in a shift of the optimal parameter $\theta_t^3$ using the TV-GP-UCB model results in frequent sampling at the acquisition bounds (dashed lines) resulting in unstable control configurations. This behavior is significantly reduced using our model \gls{uitvbo}. Including the constraints (C-\gls{uitvbo}) yields only queries very close to the optimum.}
  \label{fig:trajectory}
  \vspace{-0.3cm}
\end{figure}

\fig\ref{fig:trajectory} shows the query ``trajectory'' of the parameter $\theta_t^3$ of one run over time. We observe that after the non-stationary change, TV-GP-UCB starts to explore aggressively at the domain boundaries, resulting in several unstable controllers. While \gls{uitvbo} still yields two unstable controllers, the required exploration is significantly reduced as the model was able to adapt to the change in dynamics.
If the convexity assumption is included (green), the model can extrapolate within the bound from local queries (\cf\fig\ref{fig:constrained_GP}).
Hence, the sampling radius around the optimum is reduced, resulting in a superior tuning performance.

%% file: moin_matter/conclussion.tex
\section{Conclusion}

We proposed a modeling approach to \gls{tvbo} for controller tuning, which utilizes the concept of \gls{ui} to cope with non-stationary optimization problems. Using a \gls{lqr} tuning problem of an inverted pendulum with time-varying dynamics, we demonstrated that our modeling approach \gls{uitvbo} outperforms the state-of-the-art in terms of regret while yielding substantially fewer unstable parameter configurations.

Furthermore, we successfully merged TVBO with shape constraints to further improve performance. While the results look very promising, scaling shape constraints (such as convexity) for \glspl{gp} to higher dimensions remains an open and important research question.
Improvements therein would greatly benefit the applicability of \gls{bo} and \gls{tvbo} in controller tuning for high dimensional systems. 
Extending the presented approach using ideas from hard shape constraints as by \citet{Aubin_2020} can be a promising direction for future research.

\gls{uitvbo} can model non-stationary changes. However, it cannot predict changes or learn them from data. For future work, we plan to model drift which would allow us to predict the change in the optimizer, which is also discussed as a relevant research direction by \citet{Simonetto_2020}.